\documentclass[letterpaper, 10 pt, conference]{ieeeconf}  
\IEEEoverridecommandlockouts                              
\usepackage{graphics} 
\usepackage{color}
\usepackage{epsfig} 
\usepackage{multirow}
\usepackage{comment}
\usepackage{afterpage}
\usepackage{setspace}
\usepackage{amsmath}
\usepackage{algorithm}
\usepackage{overpic}
\usepackage[hyphens]{url}
\usepackage[noend]{algorithmic}

\title{\LARGE \bf
In-air Knotting of Rope using Dual-Arm Robot based on Deep Learning
}

\author{
Kanata Suzuki*$^{1,2}$, Momomi Kanamura*$^{2}$, Yuki Suga$^{2}$, Hiroki Mori$^{3}$ and Tetsuya Ogata$^{2,4}$
\thanks{
* The starred authors are contributed equally.
$^{1,2}$Kanata Suzuki is with Artificial Intelligence Laboratories, Fujitsu Laboratories LTD., Kanagawa 211-8588, Japan. 
$^{3}$Hiroki Mori is with Institute for AI and Robotics, Future Robotics Organization, Waseda University, Tokyo 169-8555, Japan.
$^{1,2}$Kanata Suzuki, $^{2}$Momomi Kanamura, $^{2}$Yuki Suga, and $^{2,4}$Tetsuya Ogata are with Department of Intermedia Art and Science, School of Fundamental Science and Engineering, Waseda University, Tokyo 169-8050, Japan. 
$^{2,4}$Tetsuya Ogata is also at National Institute of Advanced Industrial Science and Technology, Tokyo 100-8921, Japan. 
E-mail:{\tt\small suzuki.kanata@fujitsu.com}
}}

\begin{document}
\maketitle
\thispagestyle{empty}
\pagestyle{empty}


\begin{abstract}

In this study, we report the successful execution of in-air knotting of rope using a dual-arm two-finger robot based on deep learning.
Owing to its flexibility, the state of the rope was in constant flux during the operation of the robot. 
This required the robot control system to dynamically correspond to the state of the object at all times.
However, a manual description of appropriate robot motions corresponding to all object states is difficult to be prepared in advance.
To resolve this issue, we constructed a model that instructed the robot to perform bowknots and overhand knots based on two deep neural networks trained using the data gathered from its sensorimotor, including visual and proximity sensors. 
The resultant model was verified to be capable of predicting the appropriate robot motions based on the sensory information available online.
In addition, we designed certain task motions based on the Ian knot method using the dual-arm two-fingers robot.
The designed knotting motions do not require a dedicated workbench or robot hand, thereby enhancing the versatility of the proposed method.
Finally, experiments were performed to estimate the knotting performance of the real robot while executing overhand knots and bowknots on rope and its success rate.
The experimental results established the effectiveness and high performance of the proposed method.


\end{abstract}

\section{Introduction}
\label{sec:intro}

Manipulation of flexible objects is one of the most challenging tasks in Robotics.
This problem has been investigated by several existing studies~\cite{robot-flexible1}\cite{robot-flexible2}\cite{robot-flexible3}\cite{robot-flexible4}\cite{robot-flexible5}.
Such operations (e.g., folding clothes~\cite{robot-flexible2}\cite{dl-exp1}, folding papers~\cite{butsuri-paper}, making a bed~\cite{bed-make}, etc.) require the execution of basic motions used in daily life.
The successful manipulation of flexible objects by robots would expand their scope of application drastically.
In this study, we focus on the robotic manipulation of a rope, which is a type of linear flexible object, and realize its in-air knotting using a dual-arm two-fingers robot.

The methods of flexible object manipulation are classified into two categories, physical modeling-based approaches and machine learning-based approaches.
In physical modeling-based approaches, robotic motions are determined based on the estimated geometric information of the object. 
However, the state of the rope remains in constant flux during the knotting operation, making a complete a-priori description of all object states and their corresponding robot motions difficult to compile. 
This makes the construction of such models quite expensive~\cite{butsuri-paper}\cite{butsuri}\cite{butsuri-survey}.

On the contrary, in machine learning-based approaches, especially in those based on deep neural networks (DNNs), the sensorimotor experience of the robot can be learned~\cite{dl-exp1}\cite{dl-exp2}\cite{dl-exp3} or reinforcement learning using real robots or in a sim-to-real manner can be used~\cite{dl-rl1}\cite{dl-rl2}\cite{dl-rl3}\cite{dl-rl5}\cite{dl-rl6}.
These methods are more versatile and particularly promising for the manipulation of rope as they are constructed based on the motion data of the robot and do not depend on physical models of the target object.
Moreover, the learning methods for the sensorimotor experience of the robot~\cite{dl-exp1}\cite{dl-exp2}\cite{dl-exp3} are particularly suitable for our task because it does not require a large quantity of training data or a complicated reward design.

Versatility is also important for the manipulation of flexible objects.
Several existing studies have reported the robotic execution of knotting tasks using dedicated work environments~\cite{inaba}\cite{yamakawa} or hardware~\cite{gomi1}\cite{seo}.
However, even though they are realistic approaches for performing particular tasks, they are not versatile as it is difficult to use them for other similar tasks or environments.
We propose a model by leveraging DNNs~\cite{dl-multi1}\cite{dl-multi2} that are capable of integrating various sensor information acquired by the robot, to avoid compromising the versatility of the method.

\setlength\textfloatsep{5pt}
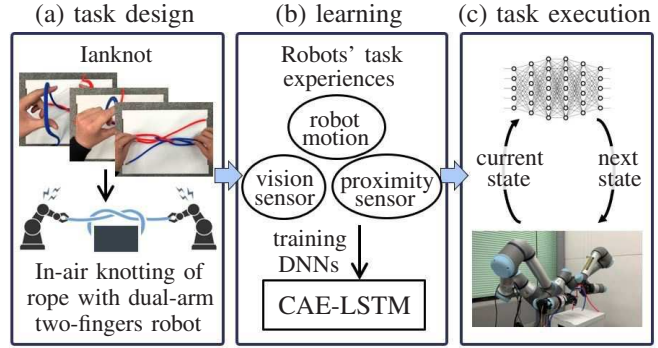
\begin{figure}[t]
    \centering
        \includegraphics[width=8.6cm]{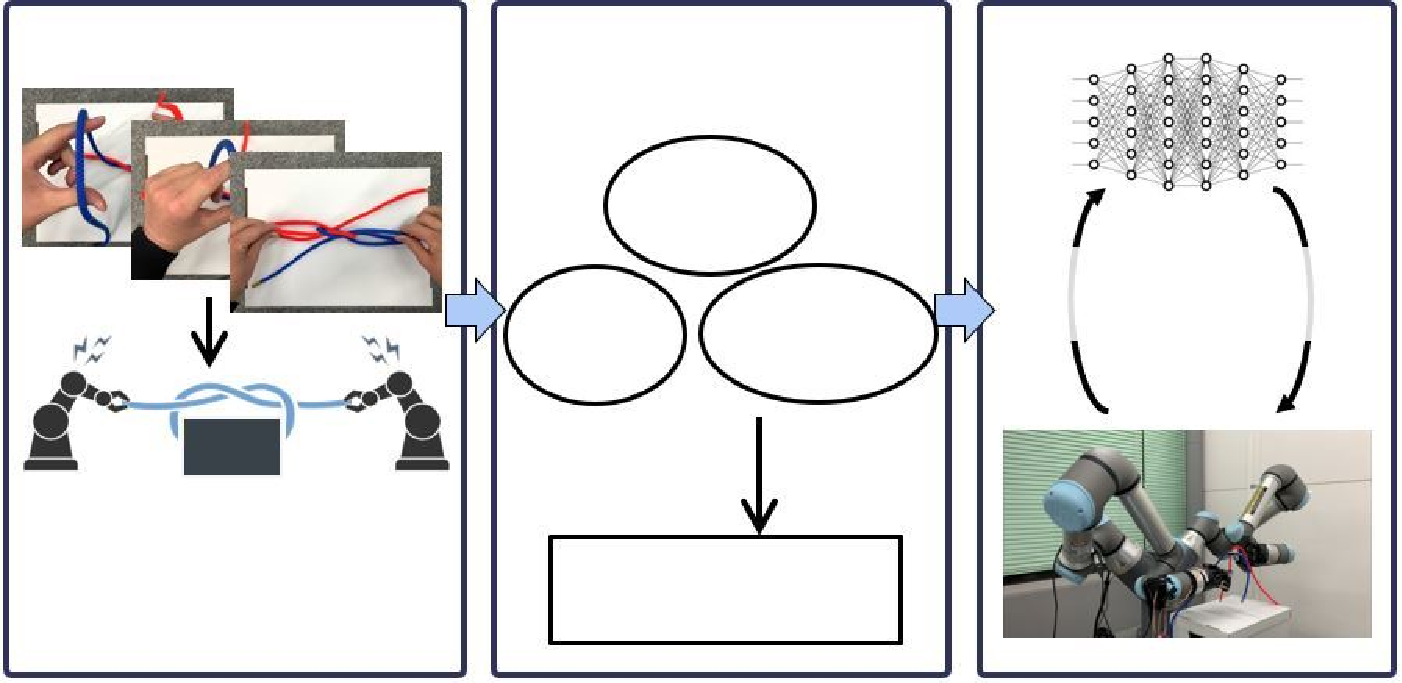}
        \put(-235.0, 123.0){\normalsize (a) task design}
        \put(-144.0, 123.0){\normalsize (b) learning}
        \put(-75.0, 123.0){\normalsize (c) task execution}
        \put(-218.0, 108.0){\small Ianknot}
        \put(-235.0, 26.0){\small In-air knotting of}
        \put(-238.0, 16.0){\small rope with dual-arm}
        \put(-235.0, 6.0){\small two-fingers robot}
        \put(-142.0, 108.0){\small Robots' task}
        \put(-141.0, 99.0){\small experiences}
        \put(-131.0, 83.5){\small robot}
        \put(-134.0, 76.5){\small motion}
        \put(-151.5, 61.0){\small vision}
        \put(-152.0, 54.0){\small sensor}
        \put(-119.5, 60.5){\small proximity}
        \put(-114.0, 53.5){\small sensor}
        \put(-145.0, 38.5){\small training}
        \put(-143.0, 28.5){\small DNNs}
        \put(-143.0, 12.5){\normalsize CAE-LSTM}
        \put(-68.0, 69.0){\small current}
        \put(-64.0, 61.0){\small state}
        \put(-22.0, 69.0){\small next}
        \put(-22.0, 61.0){\small state}
    \caption{
Overview of our study.
We adopted CAE-LSTM as the learning method for the sensorimotor experiences of the robot(see Subsection~\ref{sec:method2}).
    }
    \label{fig:intro}
\end{figure}

In this study, we propose a robot control system to execute knotting of rope based on a learning method comprising two DNNs.
Our method exhibits the following characteristics:
\begin{itemize}
    \item The designed in-air knotting motion of the robot is highly versatile (Fig.~\ref{fig:intro}(a)).
    \item A learning method is used to learn the sensorimotor experiences of the robot, including the data gathered from vision and proximity sensors (Fig.~\ref{fig:intro}(b)), and the knotting motion is generated online based on the sensory information (Fig.~\ref{fig:intro}(c)).
\end{itemize}
We designed the knotting motion of the robot based on the Ian knot method, and it does not require a dedicated workbench or robot hand.
In addition, we trained DNNs to learn the relationship between the state of the rope as surmised based on the combined sensory information and the corresponding motion of the robot.
Finally, based on experiments of real robots, we verify the capability of the proposed system to generate the appropriate knotting motion corresponding to each state of the rope online.

\section{Related Works}
\label{sec:related}

\subsection{Utilization of Input Modality in Knotting of Rope}
\label{sec:related1}

When knotting operations are classified from the viewpoint of their input modalities into two categories, the studies utilizing visual information and others utilizing sensor information other than vision.
In the former, the rope shape is visually recognized for the motion plan and control of the knotting operations.
Inaba et al.~\cite{inaba} proposed a hand-eye system to capture the position of the rope and planned robot manipulations based on this data.
Yamakawa et al.~\cite{yamakawa} executed high-speed knotting using a dynamic model of robot fingers and a rope.
Several studies have focused on topological states based on knot theory.
Matsuno et al.~\cite{matsuno} executed knotting tasks by recognizing the phase of the rope based on knot invariants.
Morita et al.~\cite{morita} constructed a model to determine appropriate knotting motions using mathematical processing based on Reidemeister moves, which are certain knot theoretic operations on projection diagrams.
Using the same theory, some studies~\cite{wakamatsu2}\cite{wakamatsu3} proposed the determination of robot motion for knotting and unraveling operations on linear flexible objects.

In studies that utilize sensor information other than vision, Gomi et al.~\cite{gomi1} performed in-air knotting of rope by utilizing the RGDB camera.
Seo et al.~\cite{seo} performed tasks by using the same robot hand in \cite{gomi1}, RGBD camera, and torque sensors.
During the knotting of rope, the rope state was in constant flux, and occlusions by the robot hand often occurred.
Thus, the robot was required to utilize various sensor information other than vision to assess the target object.
Since that information is difficult to reproduce with a physics simulator~\cite{dl-rl-sim1}\cite{dl-rl-sim2}\cite{dl-rl-sim3}, an approach to learn and generalize a robot's task experiences is preferable.

\subsection{Knotting Operations corresponding to the Rope State}
\label{sec:related2}

When the previous studies mentioned in Subsection~\ref{sec:related1} are classified from the viewpoint of the rope state during task operation into other two categories, the studies that performed knotting of rope on a workbench and in-air.
\cite{inaba}\cite{yamakawa}\cite{matsuno}\cite{morita} established the viability of knotting a rope using a single robot arm; however, they assumed the rope to be placed on a workbench.
Further, they focused purely on knotting using a robot arm.
In most practical applications, the purpose of knotting manipulation is not merely to create a knot, but to tie the rope to a target object.

\cite{gomi1}\cite{seo} realized in-air knotting of rope using a dual-arm robot.
However, in these studies, an independently developed robot hand was used, and the reproducibility of the technique was not considered.
To produce a versatile robot capable of performing multiple tasks, a robot hand designed for a specific task is not ideal.
Rather, successful realization of in-air knotting of rope using a commonly used two-finger gripper would enable the execution of different tasks using the same robot system.
In our tasks, the realization of multiple knotting techniques, such as overhand knots and bowknots using the same robot control system was desirable.

Summarizing above, the following three requirements were imposed for the knotting operation using the robot.
\begin{enumerate}
    \item Appropriate motions should be generated for all rope states and for unlearned ropes.
    \item The tasks should be executed using a dual-arm robot equipped with a commonly used robot hand.
    \item Multiple in-air knotting manipulations should be possible.
\end{enumerate}
In this study, 
to satisfy requirement 1, we utilized a robot control system consisting of two DNNs to generate the appropriate motions corresponding to the rope states based on the information obtained from visual and proximity sensors attached to the tip of the robot hand.
This enabled the capture of subtle changes that are difficult to capture using solely visual information, thereby improving the performance of the robot.
In addition, 
to satisfy requirements 2 and 3, knotting motions were designed based on the Ian knot method, and the tasks were accordingly performed using a dual-arm two-fingers robot.

\setlength\textfloatsep{5pt}
\begin{figure}[t]
    \centering
        \includegraphics[height=5.4cm,width=8.6cm]{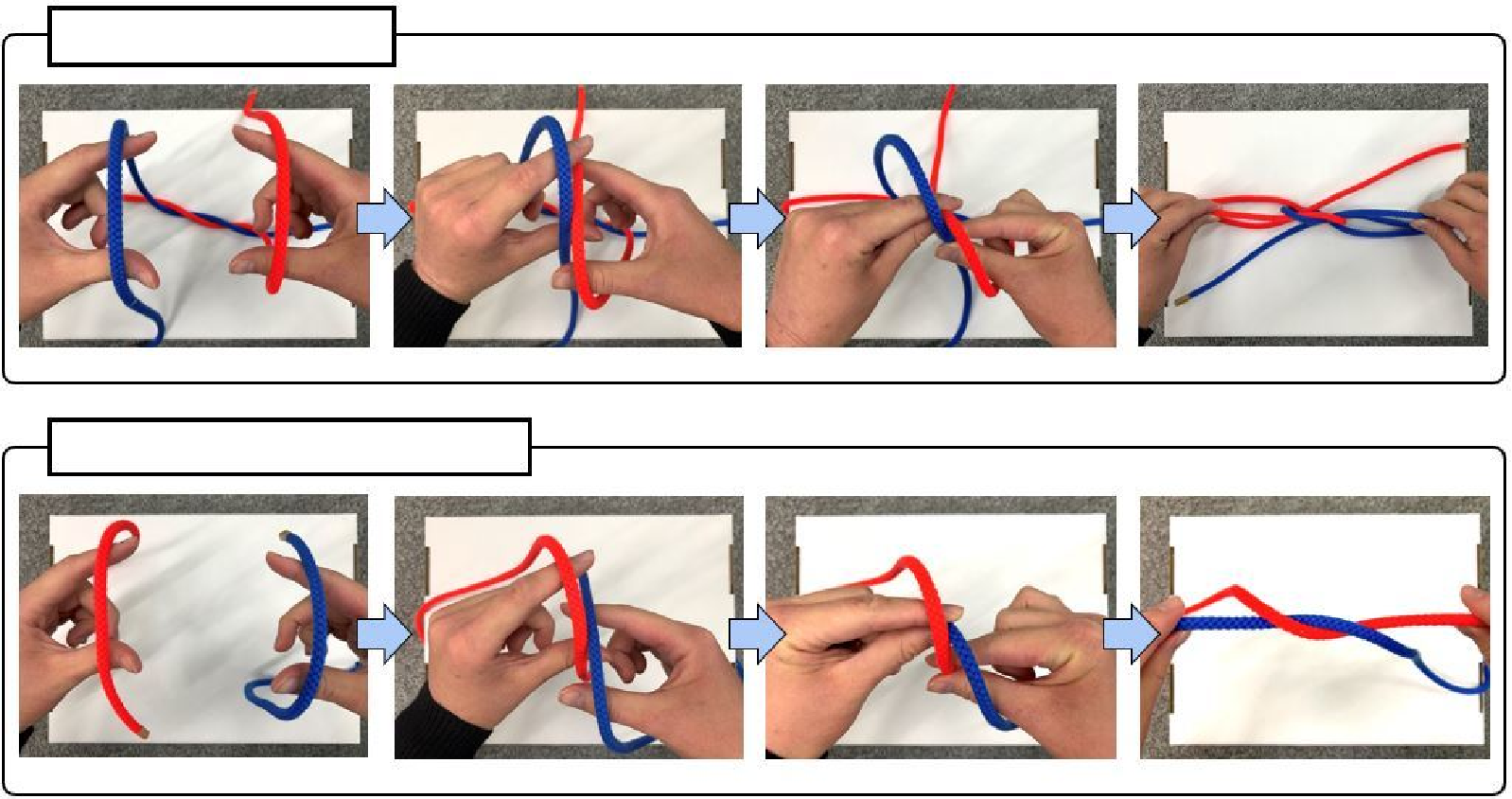}
        \put(-233.0, 143.3){\small (a) Bowknot}
        \put(-233.0, 64.4){\small (b) Overhand knot}
    \caption{
Examples of the Ian knot method.
Both the (a) bow knot and (b) overhand knot can be executed using the same procedure.
  }
  \label{fig:ianknot}
\end{figure}

\setlength\textfloatsep{5pt}
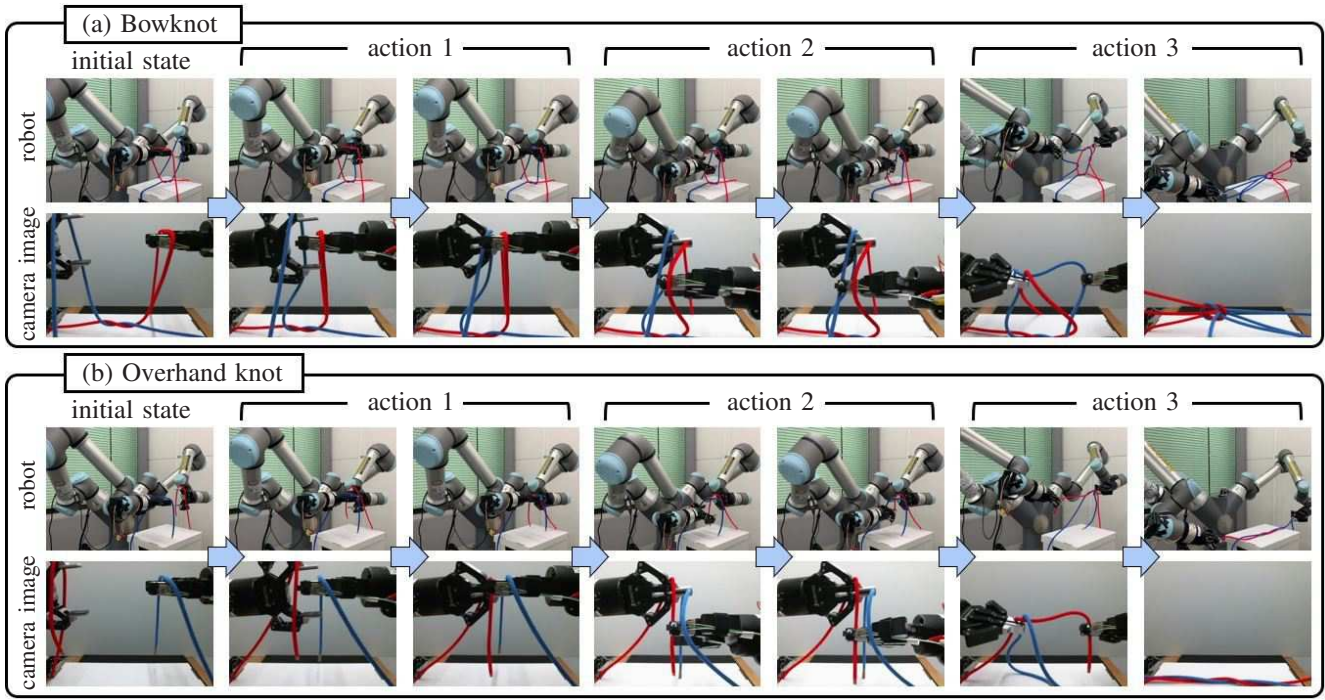
\begin{figure*}[t]
    \centering
        \includegraphics[width=17.6cm]{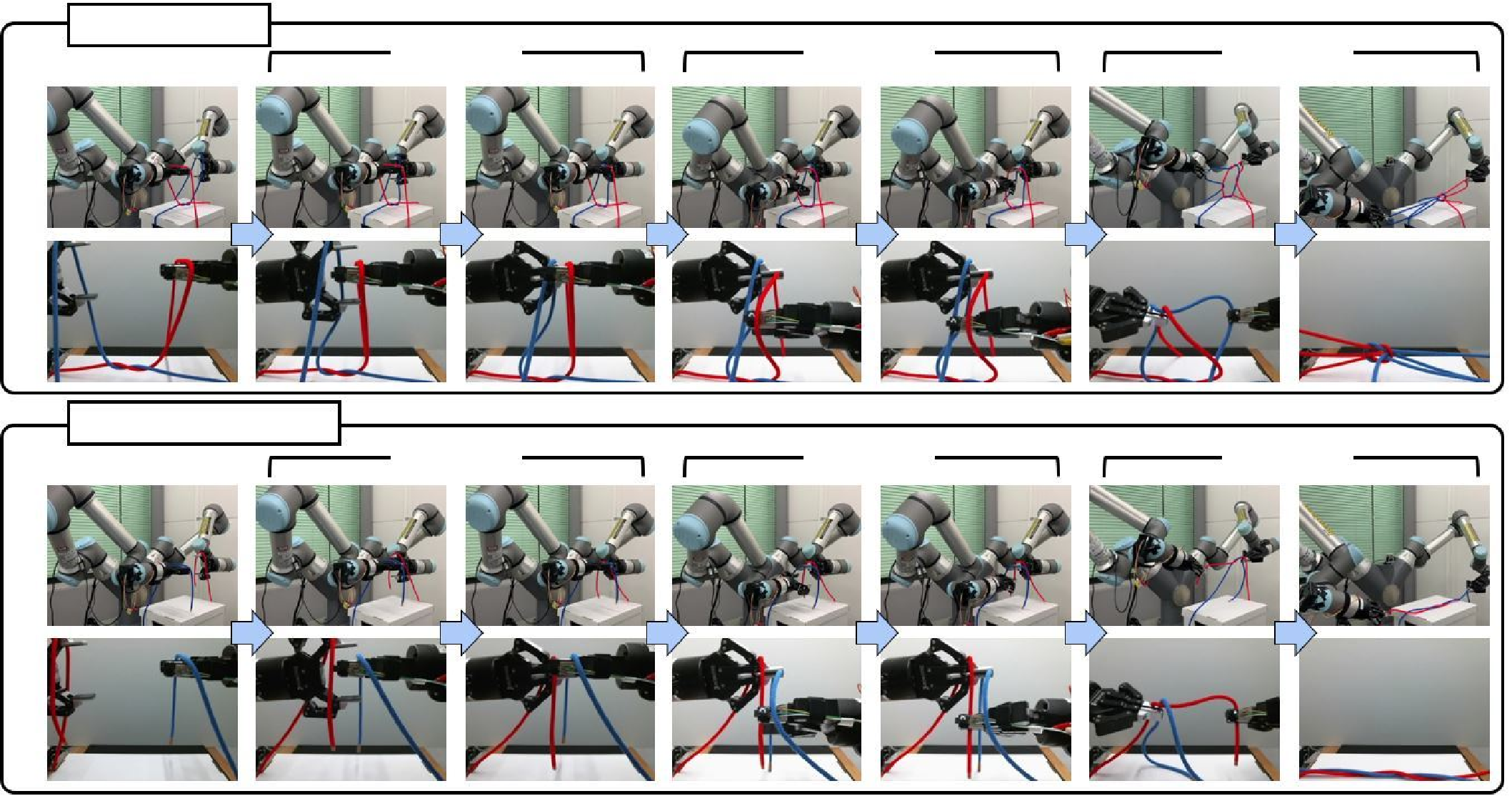}
        \put(-472.0, 254.0){\normalsize (a) Bowknot}
        \put(-476.0, 240.0){\normalsize initial state}
        \put(-364.0, 245.0){\normalsize action 1}
        \put(-228.0, 245.0){\normalsize action 2}
        \put(-90.0,  245.0){\normalsize action 3}
        \put(-472.0, 122.0){\normalsize (b) Overhand knot}
        \put(-476.0, 108.0){\normalsize initial state}
        \put(-364.0, 111.0){\normalsize action 1}
        \put(-228.0, 111.0){\normalsize action 2}
        \put(-90.0,  111.0){\normalsize action 3}
        \put(-496.0, 205.0) {\small \rotatebox{90}{robot}}
        \put(-496.0, 137.0) {\small \rotatebox{90}{camera image}}
        \put(-496.0, 73.0) {\small \rotatebox{90}{robot}}
        \put(-496.0, 5.0) {\small \rotatebox{90}{camera image}}
    \caption{
Hand-crafted in-air knotting motions for (a) bowknot and (b) overhand knot using a dual-arm two-fingers robot.
    }
    \label{fig:knot-robot}
\end{figure*}

\section{Method}
\label{sec:method}

\subsection{Design of Task Motion}
\label{sec:method1}

As described in Subsection~\ref{sec:related2}, the realization of multiple knotting operations using a single robot is a central goal of this study.
Most knotting operations performed by humans require the use of three or more fingers, making them difficult to execute using a two-finger gripper.
In this study, we focus on a knotting technique called the Ian knot, which can be used to realize a bowknot and an overhand knot with fewer steps than traditional methods~\cite{ianknot}.
Further, it enables the realization of both types of knots using the same procedure (see Fig.~\ref{fig:ianknot}).
This enables the realization of in-air bowknots and overhand knots without requiring complicated motions from the robot, such as swapping hands to grab the rope.

Figure~\ref{fig:knot-robot} depicts the motions required to realize in-air bowknots and overhand knots using a dual-arm robot.
Each knotting motion was hand-crafted, and they executed by controlling the robot's joint angles to key postures.
The task operation can be decomposed into the following elementary actions and robot states.
\begin{itemize}
    \item Twisting the rope symmetrically, and hanging it on the fingers of both hands (initial state).
    \item Crossing the left and right ropes, and grabbing the lying rope using the left gripper (action 1).
    \item Moving until the fingertip of the right gripper touches the vertically hanging rope, and grabbing it using the center of the finger (action 2).
    \item Pulling both arms apart, and fastening the rope (action 3).
\end{itemize}
The motions to execute bowknots and overhand knots can be distinguished based on the rope states during the robot's actions.
Especially in the case of action 2, the robot needs to grab the hanging rope to which an external force is applied, requiring the motion to be adjusted to the rope state.

In the Ian knot method, both arms are required to be brought close to each other during the task.
However, in a dual-arm robot, a collision between hands is to be avoided to prevent damage to the hardware.
Therefore, we used a proximity sensor during the execution of action 2 and designed safe operations enabling the robot to grab the rope with one of its hands and then the other.
The distance to the object in front of the proximity sensor was estimated using it and the appropriate task motions were selected based on the value.
Thus, besides the visual information, the observation made by the proximity sensor also played an important role in recognizing the rope state.

\setlength\textfloatsep{5pt}
\begin{figure}[t]
    \centering
        \includegraphics[width=8.6cm]{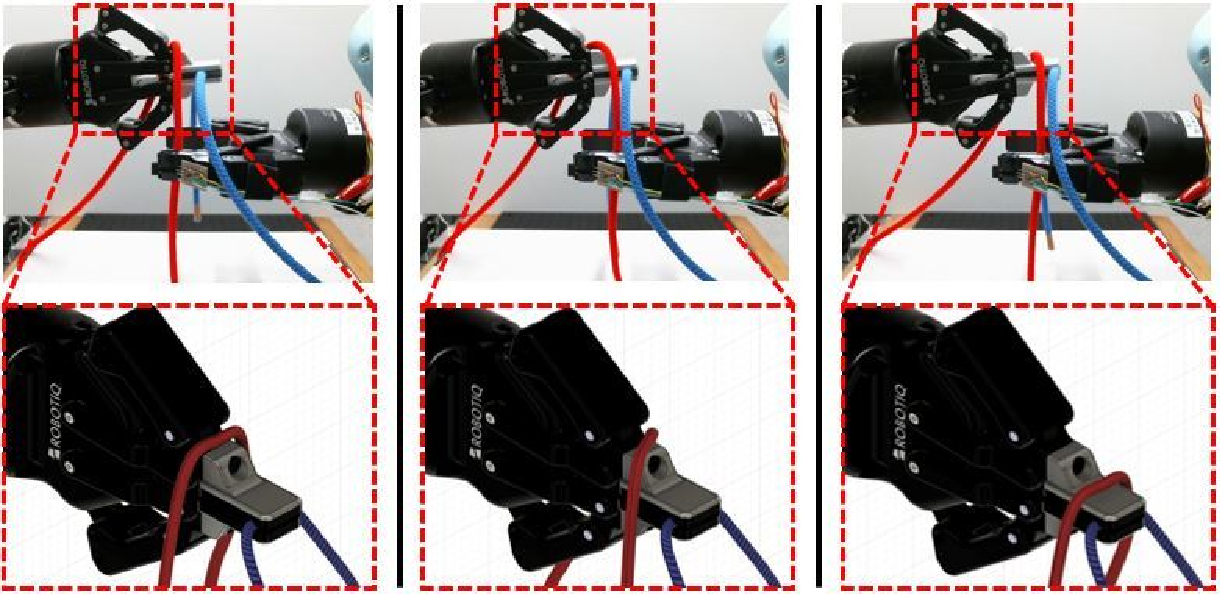}
        \put(-229.0, -10.0){\small (a) Pattern 1}
        \put(-147.0, -10.0){\small (b) Pattern 2}
        \put(-63.0, -10.0){\small (c) Pattern 3}
    \caption{
Patterns of the rope state in overhand knot operation.
  }
  \label{fig:rope-kakari}
\end{figure}

And in the overhand knot operation, the degree of hooking the rope to the left gripper in the initial state of the task greatly affected the subsequent robot motion.
This was observed to be a phenomenon peculiar to overhand knots in which the rope to be grabbed in action 2 was not fixed.
We classified the rope state into three patterns: the rope was hung parallelly from the base of the gripper's finger (Fig.~\ref{fig:rope-kakari}(a)), diagonally from the base of the finger (Fig.~\ref{fig:rope-kakari}(b)), and parallelly from the tip of the finger (Fig.~\ref{fig:rope-kakari}(c)).
In such cases, the robot was required to move its right gripper to an appropriate grasping position.
That is a problem peculiar to knotting operation that is difficult to perform by the hand-crafted method, and experiment 2 in Subsection~\ref{sec:result2} showed the effectiveness of the proposed learning method in comparison with the hand-crafted method.

\subsection{Learning Method: CAE-LSTM}
\label{sec:method2}

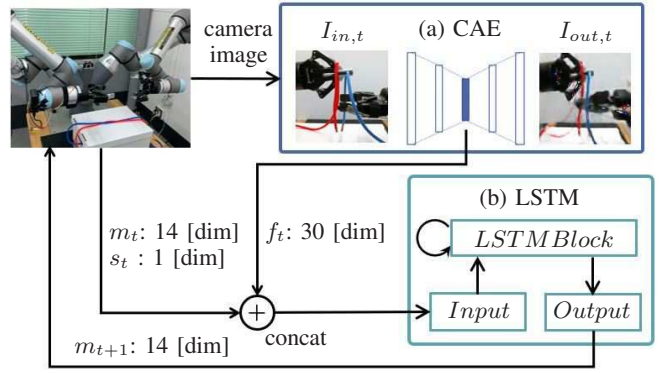
\begin{figure}[t]
  \centering
        \includegraphics[width=8.6cm]{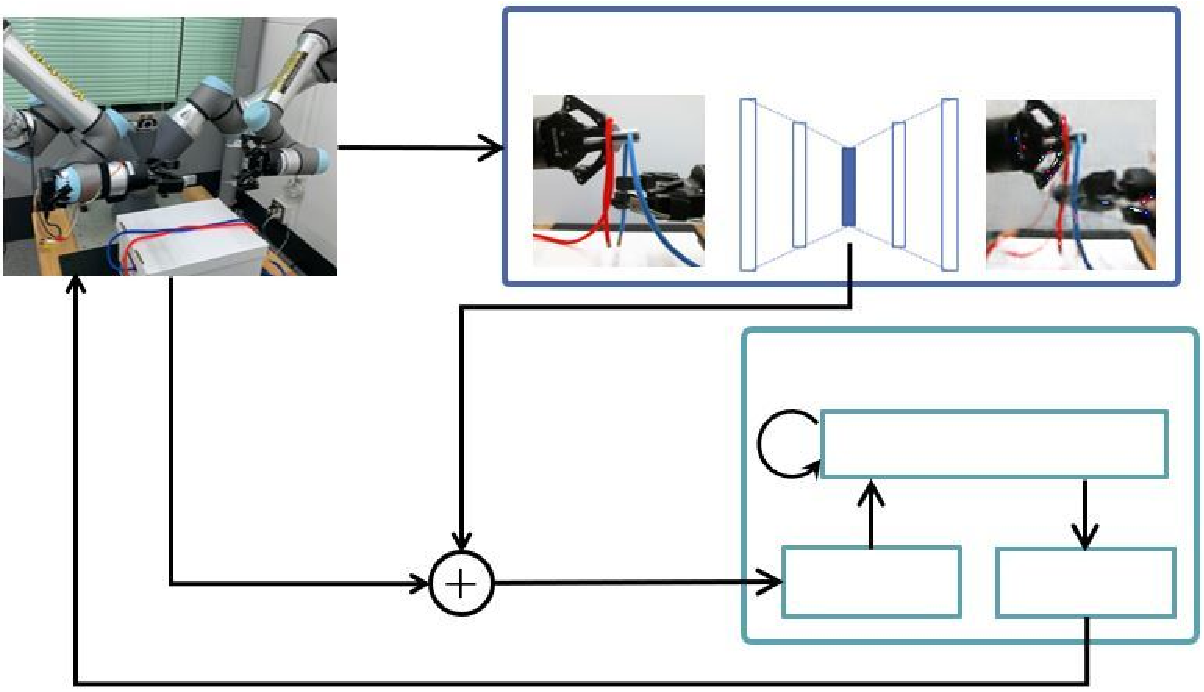}
        \put(-171.0, 127.0){\small camera}
        \put(-169.0, 117.0){\small image}
        \put(-90.0,  127.0){\small (a) CAE}
        \put(-127.0, 126.0){\small $I_{in, t}$}
        \put(-37.0,  126.0){\small $I_{out, t}$}
        \put(-67.0, 63.0){\small (b) LSTM}
        \put(-69.0, 47.0){\small $LSTM Block$}
        \put(-79.0, 20.0){\small $Input$}
        \put(-39.0, 20.0){\small $Output$}
        \put(-148.0, 10.0){\small concat}
        \put(-148.0, 50.0){\small $f_t$: 30 [dim]}
        \put(-207.0, 50.0){\small $m_t$: 14 [dim]}
        \put(-207.0, 40.0){\small $s_t$ : 1 [dim]}
        \put(-220.0, 7.0){\small $m_{t+1}$: 14 [dim]}
  \caption{
Overview of the proposed learning method comprising two DNNs, (a) CAE and (b) LSTM.
  }
  \label{fig:method}
\end{figure}

We constructed a robot control model by learning the robot motions described in the previous subsection.
The proposed learning method consisted of two DNNs, Convolutional Auto-Encoder (CAE) and Long short-term memory (LSTM).
Therefore, the proposed method was named CAE-LSTM.
It included the following two stages of training.
First, CAE was trained using the acquired image data to extract image features that are significant for the experiences of the sensorimotor (Fig.~\ref{fig:method}(a)).
Subsequently, LSTM was trained using time-series data composed of image features, proximity sensor information, and robot joint angles (Fig.~\ref{fig:method}(b)).
By learning to integrate sensorimotor experiences, including the data gathered from multiple sensors, and motion information using the LSTM, interactive motions were generated corresponding to the varying rope state.
Details of the training dataset and the model parameters are described in Section~\ref{sec:exp}.

\subsubsection{CAE}
To learn the information gathered from multiple sensors and pertaining to multiple motions, the extraction of important features from high-dimensional sensor information was necessary.
To this end, we used CAE, a DNN capable of extracting image features from high-dimensional raw images.
The CAE comprised an encoder and decoder.
The encoder consisted of convolutional layers and fully-connected layers, and the decoder consisted of deconvolutional layers and fully-connected layers (see Table I).
The output of the CAE was calculated as follows:
\begin{eqnarray}
    f = Enc(I_{in}), \\
    I_{out} = Dec(f),
\end{eqnarray}
where, $Enc$ denotes the encoder of the CAE, $Dec$ denotes the decoder of the CAE, $f$ denotes the image feature vector, $I_{in}$ denotes the input image, and $I_{out}$ denotes the output image of the CAE.
During the training phase, the parameters of the CAE were optimized to provide output values equal to the input values based on the following loss function, $L_{ae}$:
\begin{eqnarray}
    L_{ae} = \frac{1}{N} \sum_n^{N} {(I_{in, n} - I_{out, n})^2},
\end{eqnarray}
where, $N$ denotes the total number of images in the training dataset.
Successful reconstruction of the input image by the CAE indicated that the encoded low-dimensional vectors reflected the relationship between the robot manipulator and object states faithfully.
The image features extracted by the trained CAE were used to train LSTM, as described in the following subsection.

\setlength\textfloatsep{5pt}

\subsubsection{LSTM}
We trained the LSTM using image features, the output of the proximity sensor, and the joint angles of the robot.
The LSTM is a recurrent neural network capable of learning long-term dependencies in time-series data and has been used in multiple studies on robot motion learning~\cite{kase}\cite{lstm-robot1}\cite{lstm-robot2}.
We used LSTM as a robot control system to predict the target robot posture during the following step.
The LSTM output the next state of the sequence based on the previous state and the hidden and cell states, using the following equation:
\begin{eqnarray}
    y_{t+1}, h_{t+1}, c_{t+1} = LSTMBlock(m_t, f_t, s_t, h_t, c_t), \\
    m_{t+1}, f_{t+1}, s_{t+1} = tanh(FC(y_{t+1})),
\end{eqnarray}
where, $m$ denotes the robot joint angle, $f$ denotes the collection of image features, $s$ denotes the value of the proximity sensor, $h$ and $c$ denote the hidden and cell states of the LSTM, $t$ denotes the current timestep, and $\bullet_t$ denotes the value at the timestep, $t$.
Further, $tanh$ is an activation function, $LSTMBlock$ denotes an LSTM layer consisting of multiple gates, e.g., forgetting gates, $y$ denotes the output of the LSTM layer, and $FC$ denotes a fully-connected layer.
The loss function of the LSTM was defined as follows:
\begin{eqnarray}
    L_{rnn} = \sum_{t=0}^{T-1} \left[ (m_{t+1}-\hat{m}_{t+1})^2 
                                    + (f_{t+1}-\hat{f}_{t+1})^2 \right. \nonumber \\
                                     \quad \left. + (s_{t+1}-\hat{s}_{t+1})^2 \right],
\end{eqnarray}
where, $T$ denotes the sequence length, and $\hat{\bullet}$ denotes the target value obtained from the training dataset.
Since the LSTM learns information obtained from multiple sensors and pertaining to multiple motions equally, the model embedded the relationship between the robot's motion and the state of the target object surmised based on the sensor information.

\subsection{Online Motion Generation}
\label{sec:method3}

During the motion generation phase, the data acquired from the current robot state was used as the input to the CAE and the LSTM.
The initial value of the hidden state, $h_0$, was set to 0.
The joint angles, $m_{t+1}$, predicted by the LSTM were set to define the target postures of the robot, and each joint was appropriately controlled.
The system repeated the aforementioned process online.

\section{Experiment}
\label{sec:exp}

\subsection{Hardware}
\label{sec:task}

\begin{figure}[t]
    \centering
        \includegraphics[width=8.6cm]{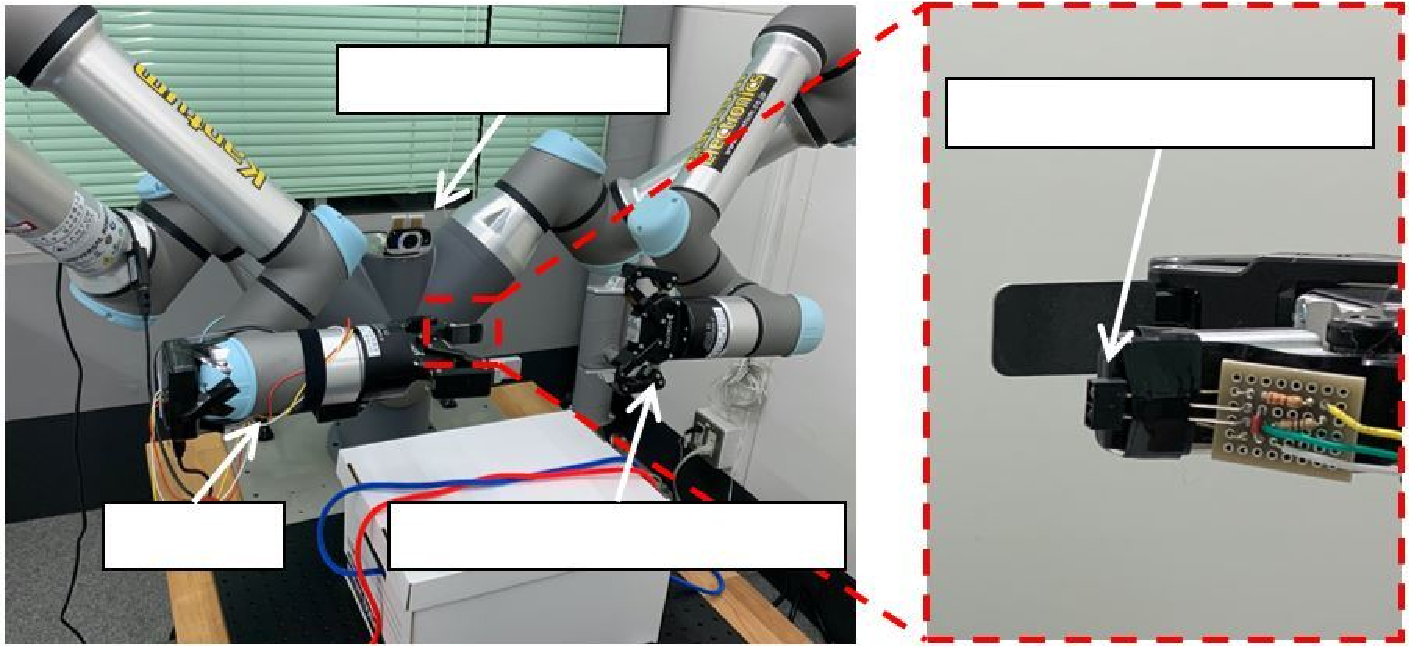}
        \put(-181.0, 95.2){\small RGB camera}
        \put(-222.0, 16.0){\small UR5e}
        \put(-174.0, 16.5){\small Two-fingers gripper}
        \put(-74.5, 89.4){\small Proximity sensor}
    \caption{
The experimental environment involving the dual-arm robot.
    }
    \label{fig:exp}
\end{figure}

The experimental environment is depicted in Fig.~\ref{fig:exp}.
In our experiments, the target rope was knotted to a box by the robot.
We used a dual-arm robot consisting of two UR5e attached to a Y-shaped body.
A general two-finger gripper~\cite{robotiq} was attached to each arm.
A camera was mounted on the body to enable oversight over the operations of the robot hand.
A photo reflector was used as a proximity sensor to detect the rope.
It was a reflective photo interrupter capable of estimating its distance to an object by reflecting light onto the target object.
A detection distance of 10 [mm] was used for the photo reflector used in this study and its output value was incorporated within the input of the LSTM.
Our robot systems were built on RT-Middleware-based systems~\cite{sensors}.

\begin{figure}[t]
    \begin{minipage}{0.48\hsize}
        \begin{center}
            \includegraphics[width=4.2cm]{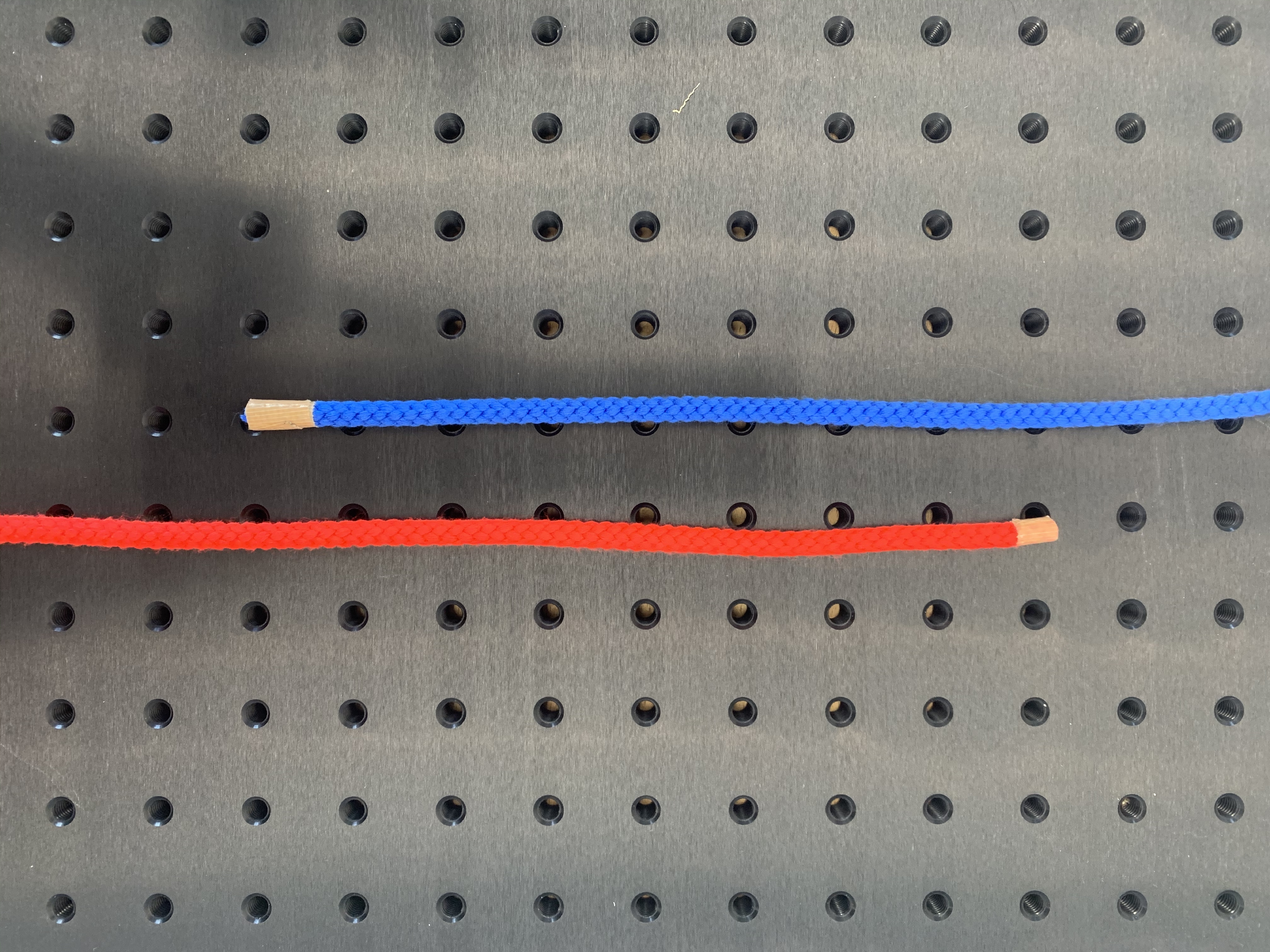}
        \end{center}
        \caption{The rope used during the training task sequences.}
        \label{fig:rope-train}
    \end{minipage}
    \begin{minipage}{0.48\hsize}
        \begin{center}
            \includegraphics[width=4.2cm]{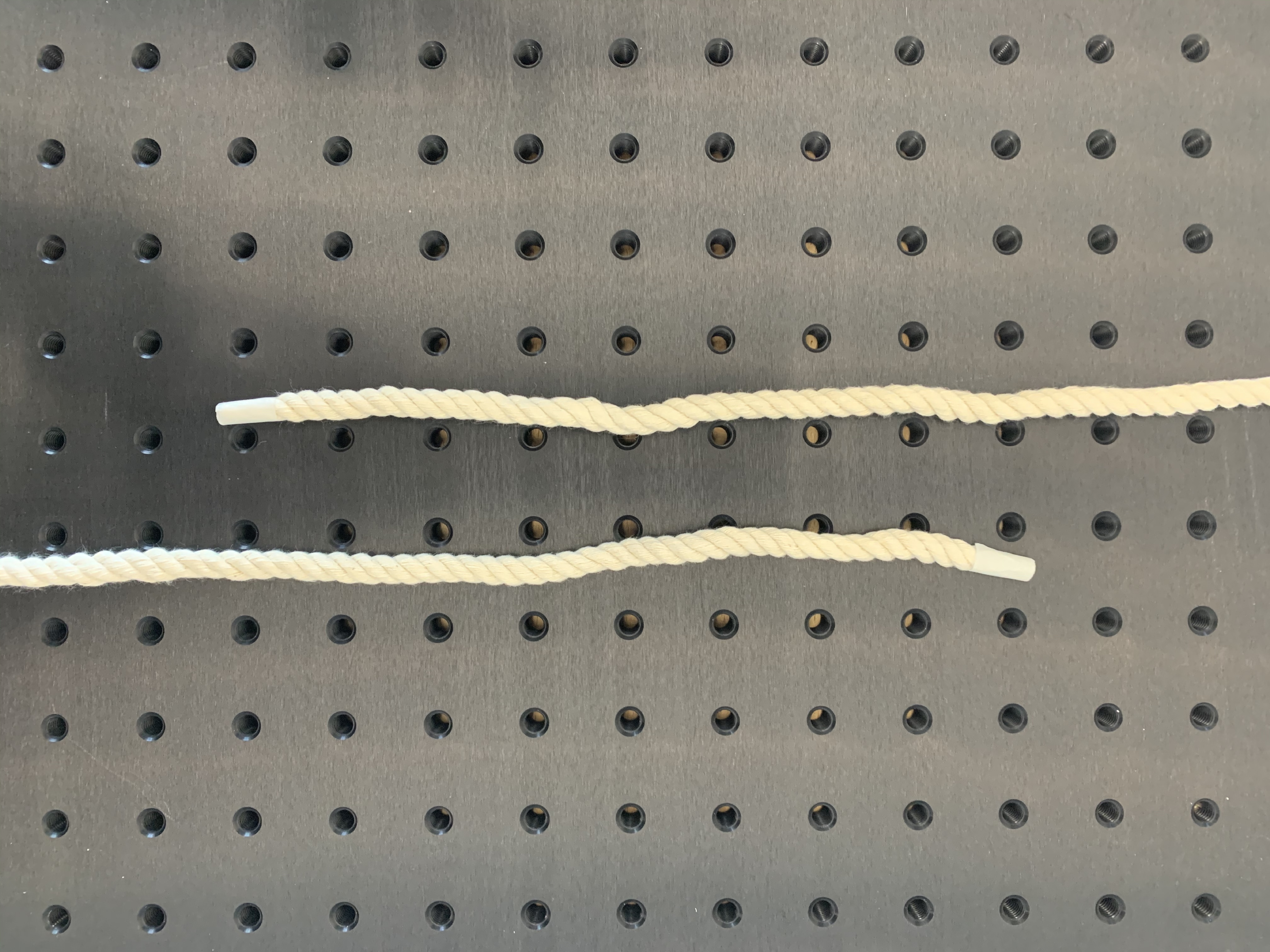}
        \end{center}
        \caption{Unlearned rope used in experiment 3.}
        \label{fig:rope-test}
    \end{minipage}
\end{figure}

For the object to be manipulated, we used an acrylic rope with a width of 5 [mm] (Fig.~\ref{fig:rope-train}).
The color of the rope was red and blue, which changed near the center.
The rope was placed such that the red part was visible on the left of the box and the blue part was visible on the right side.
The experimental setup was based on that of a previous study~\cite{wang}, which proposed a reduction in the cost of recognizing the state of the rope based on its color.
Although we used dedicated rope, the generalization performance of the method proposed in this study was confirmed by using an unlearned object in experiment 3 which is described in the next subsection.
Based on the aforementioned implementation, the robot executed the bowknot and overhand knot operations.

\setlength\textfloatsep{5pt}

\setlength\textfloatsep{5pt}

\begin{table}[t]
    \centering
    \begin{tabular}{c|c}
        \multicolumn{2}{c}{TABLE I: Structures of the DNNs} \\
        \hline
        Network      & Dims \\
        \hline \hline
        CAE$^{1,2}$  & input@3chs - convs@(32-64-128)chs - \\ 
                     & full@1000 - full@30 - full@1000 - \\ 
                     & dconvs@(128-64-32)chs - output@3chs \\ 
        \hline
        LSTM         & $IO$@45 - $hidden/cell$@250 (1 layer) \\ 
        \hline 
    \end{tabular}
    \begin{flushleft}
        1) all conv and dconv are kernel 6, stride 2, and padding 1. \\
        2) all layers have batch normalization and relu activation. \\
    \end{flushleft}
\end{table}

\subsection{Performance Evaluation}
\label{sec:exp-setup}

We verified the effectiveness of the proposed method based on three kinds of experiments.
In each experiment, we adjudged the trial in which a knot was confirmed on the rope after the task to be successful.

\textbf{experiment 1)}
In this experiment, we verified the capability of the proposed method to learn multiple knotting methods simultaneously.
We prepared 15 task sequences for bowknot and overhand knotting operations, and a total of 30 sequences were used as training data.
The rope state during the overhand knot operation was taken to be pattern 1 in this experiment.
During the execution of tasks, the robot was not given any explicit instruction on which knotting operation to perform. 
Thus, the model needed to select the appropriate knotting operation based on 
the visual sensor.

\textbf{experiment 2)}
In the overhand knot operation, the degree of hooking the rope to the left gripper in the initial state of the task greatly affected the subsequent robot motion.
We verified the capability of the proposed method to generate the appropriate motion corresponding to the rope state described in Subsection~\ref{sec:method1} and Fig.~\ref{fig:rope-kakari}.
We prepared 15 task sequences corresponding to each pattern of the rope state, and a total of 45 sequences were used as training data.
The model was required to detect subtle changes in the rope state, which were difficult to be captured based solely on visual information, and perform appropriate motions.

\textbf{experiment 3)}
In this experiment, we verified the capability of our method to perform tasks with an unlearned object.
We used a white rope (Fig.~\ref{fig:rope-test}) as an unlearned object.
The verifications of experiments 1 and 2 were conducted again by using the same parameters of the trained model as in the case of the red and blue rope.
To perform the tasks, it was necessary to generalize the position of the rope using the CAE and utilize the proximity sensor information.

\subsection{Training Setup}
\label{sec:training}

The training data used in experiments 1 to 3 were created by performing hand-crafted knotting operations that transitioned the robot posture to the key postures of the task motion in the correct sequence.
The training data for images, the output of the proximity sensor, and joint angles were sampled at five frames per second.

The detailed parameters of the models are shown in Table I.
The CAE was trained to extract 30-dimensional image features from $128\times 128\times 3$ [pixel] raw images.
The LSTM had 45 input/output neurons (image features: 30 [dim], joint angles and gripper of each arm: $7\times2=14$ [dim], and proximity sensor: 1 [dim])
We trained the CAE for 300 epochs, and the LSTM for 20000 epochs.
The batch size during the training of CAE was taken to be 24.
The RNN input was scaled to [$-$1.0, 1.0].
During the training of CAE and LSTM, we used an Adam optimizer~\cite{adam}.
The parameters of the optimizer were set to $\alpha=0.001$, $\beta_1=0.9$, and $\beta_2=0.999$.
We augmented the task sequences by adding Gaussian noise and applying color augmentation to increase the robustness of CAE and LSTM.

\begin{table*}[t]
    \centering
    \begin{tabular}{c|c|cc}
        \multicolumn{4}{c}{TABLE II: Task success rates in experiments 1 and 3} \\
        \hline
                        & \multirow{2}{*}{Object} & \multicolumn{2}{c}{Type of knotting motion} \\
        \cline{3-4}
                        &                & Bowknot        & Overhand knot \\
        \hline \hline
        \multirow{2}{*}{CAE-LSTM (ours)} 
                        & red and blue rope   & 95.0\% (19/20) & 95.0\% (19/20) \\
        \cline{2-4}
                        & white rope (unlearned) & 90.0\% (18/20) & 85.0\% (17/20) \\
        \hline
    \end{tabular}
\end{table*}

\begin{table*}[t]
    \centering
    \begin{tabular}{c|c|ccc}
        \multicolumn{5}{c}{TABLE III: Task success rates in experiments 2 and 3} \\
        \hline
                        & \multirow{2}{*}{Object} & \multicolumn{3}{c}{Rope state} \\
        \cline{3-5}
                        &                         & Pattern 1        & Pattern 2      & Pattern 3 \\
        \hline \hline
        Hand-crafted knotting operation & red and blue rope & 23.8\% (15/63) & 30.6\% (15/49) & 37.5\% (15/40) \\
        \hline
        \multirow{2}{*}{CAE-LSTM (ours)} 
                        & red and blue rope     & 95.0\% (19/20) & 95.0\% (19/20) & 100.0\% (20/20) \\
        \cline{2-5}
                        & white rope (unlearned)   & 90.0\% (18/20) & 90.0\% (18/20) & 85.0\% (17/20) \\
        \hline
    \end{tabular} 
\end{table*}

\setlength\textfloatsep{5pt}
\begin{figure}[t]
    \begin{minipage}{0.48\hsize}
        \begin{center}
            \includegraphics[width=\columnwidth]{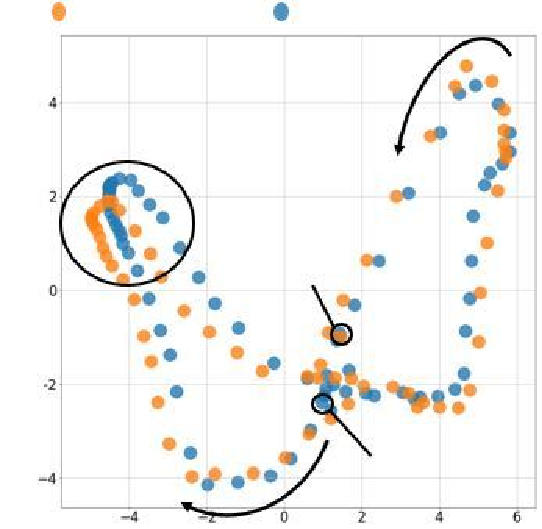}
            \put(-117.0, 50.0) {\small \rotatebox{90}{PC3}}
            \put(-60.0, -10.0) {\small PC1}
            \put(-100.0, 110.5) {\footnotesize Bowknot}
            \put(-52.0, 110.5) {\footnotesize Overhand knot}
            \put(-97.0, 84.0) {\footnotesize action 2}
            \put(-58.0, 63.0) {\footnotesize Start}
            \put(-58.0, 55.0) {\footnotesize state}
            \put(-36.0, 17.0) {\footnotesize End}
            \put(-36.0, 9.0) {\footnotesize state}
        \end{center}
        \caption{
Internal LSTM state of generated motions in experiment 1 (PC1:24.3\%, PC3:9.0\%).
}
        \label{fig:pca1-1}
    \end{minipage}
    \begin{minipage}{0.48\hsize}
        \begin{center}
            \includegraphics[width=\columnwidth]{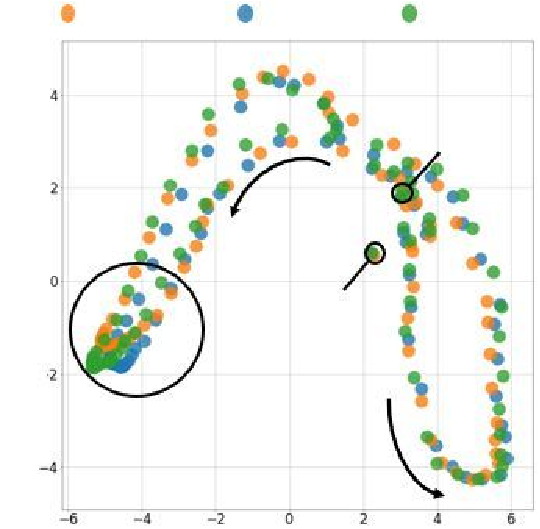}
            \put(-117.0, 50.0) {\small \rotatebox{90}{PC3}}
            \put(-60.0, -10.0) {\small PC1}
            \put(-98.5, 110.5) {\footnotesize Patt. 1}
            \put(-60.5, 110.5) {\footnotesize Patt. 2}
            \put(-24.5, 110.5) {\footnotesize Patt. 3}
            \put(-88.0, 20.0) {\footnotesize action 2}
            \put(-55.0, 46.0) {\footnotesize Start}
            \put(-55.0, 38.0) {\footnotesize state}
            \put(-21.0, 92.0) {\footnotesize End}
            \put(-21.0, 84.0) {\footnotesize state}
        \end{center}
        \caption{
Internal LSTM state of generated motions in experiment 2 (PC1:28.3\%, PC3:9.3\%).
}
        \label{fig:pca1-2}
    \end{minipage}
\end{figure}

\setlength\textfloatsep{5pt}
\begin{figure*}[t]
    \centering
        \includegraphics[width=17.6cm]{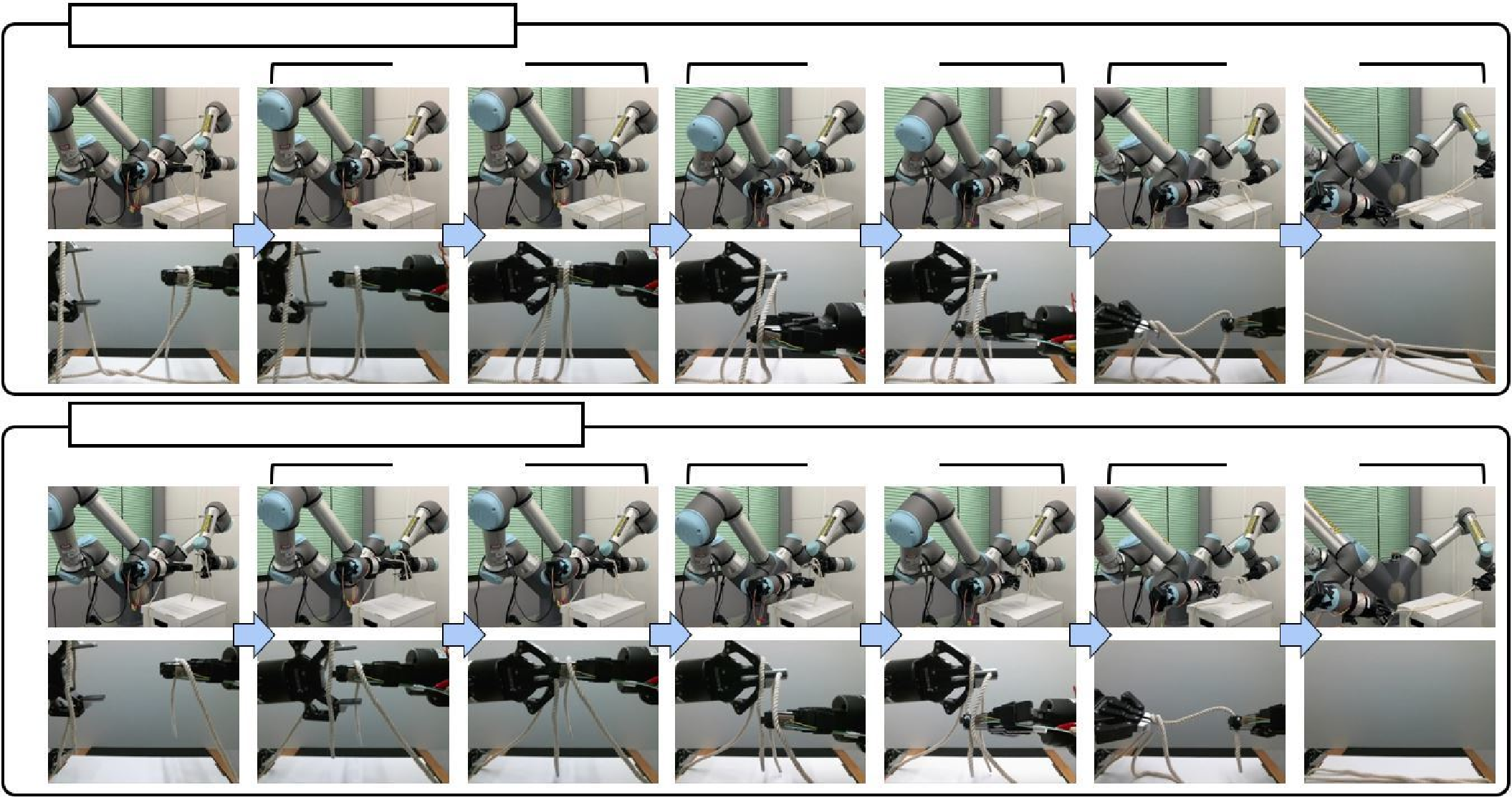}
        \put(-472.0, 254.0){\normalsize (a) Bowknot with unlearned rope}
        \put(-476.0, 240.0){\normalsize initial state}
        \put(-364.0, 241.0){\normalsize action 1}
        \put(-228.0, 241.0){\normalsize action 2}
        \put(-90.0,  241.0){\normalsize action 3}
        \put(-472.0, 122.0){\normalsize (b) Overhand knot with unlearned rope}
        \put(-476.0, 108.0){\normalsize initial state}
        \put(-364.0, 108.0){\normalsize action 1}
        \put(-228.0, 108.0){\normalsize action 2}
        \put(-90.0,  108.0){\normalsize action 3}
        \put(-496.0, 205.0) {\small \rotatebox{90}{robot}}
        \put(-496.0, 137.0) {\small \rotatebox{90}{camera image}}
        \put(-496.0, 73.0) {\small \rotatebox{90}{robot}}
        \put(-496.0, 5.0) {\small \rotatebox{90}{camera image}}
    \caption{
Examples of generated (a) bowknot and (b) overhand knot motions with unlearned rope.
    }
    \label{fig:result}
\end{figure*}

\setlength\textfloatsep{5pt}
\begin{figure}[t]
    \begin{minipage}{0.48\hsize}
        \begin{center}
            \includegraphics[width=\columnwidth]{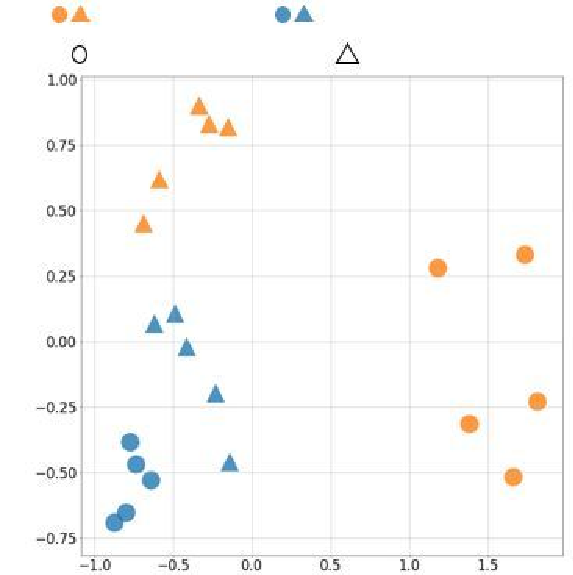}
            \put(-117.0, 50.0) {\small \rotatebox{90}{PC2}}
            \put(-60.0,  -7.0) {\small PC1}
            \put(-98.0, 113.0) {\footnotesize Bowknot}
            \put(-52.0, 113.0) {\footnotesize Overhand knot}
            \put(-98.0, 105.0) {\footnotesize trained rope}
            \put(-43.0, 105.0) {\footnotesize test rope}
        \end{center}
        \caption{
Averaged internal LSTM state of generated bowknot and overhand knot motions (PC1:58.4\%, PC2:14.9\%).
}
        \label{fig:pca2-1}
    \end{minipage}
    \begin{minipage}{0.48\hsize}
        \begin{center}
            \includegraphics[width=\columnwidth]{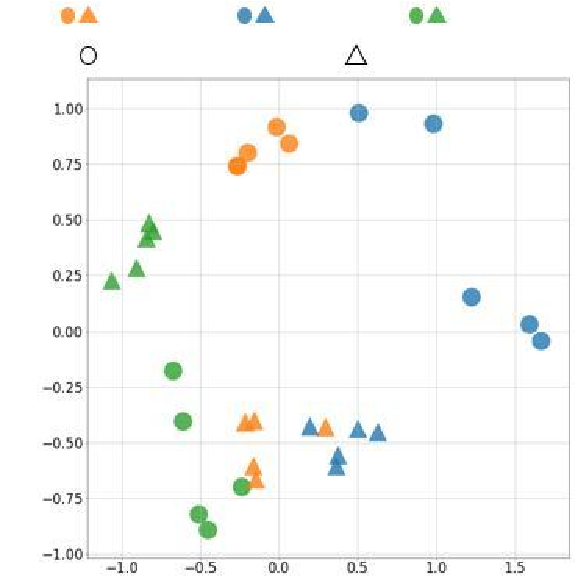}
            \put(-117.0, 50.0) {\small \rotatebox{90}{PC2}}
            \put(-60.0,  -7.0) {\small PC1}
            \put(-96.0,  113.0) {\footnotesize Patt. 1}
            \put(-60.0,  113.0) {\footnotesize Patt. 2}
            \put(-25.0,  113.0) {\footnotesize Patt. 3}
            \put(-96.0,  105.0) {\footnotesize trained rope}
            \put(-41.0,  105.0) {\footnotesize test rope}
        \end{center}
        \caption{
Averaged internal LSTM state of generated overhand knot motions for patterns of rope state (PC1:32.3\%, PC2:22.2\%).
}
        \label{fig:pca2-2}
    \end{minipage}
\end{figure}

\section{Result and Discussion}

\subsection{Experiment 1: Learning Multiple Knotting Operations}
\label{sec:result1}

First, we verified the capability of the proposed method to learn multiple knotting methods.
The model was trained using the task sequences of overhand knot and bowknot operations.
Each knotting operation was tested 20 times using online motion generation.
As depicted in the first row of Table II, the success rate was observed to be 95.0 \% for overhand knots and 95.0 \% for bow knots.
Thus, the proposed method exhibited sufficient task performance.
The failure cases were attributed to the inability of the robot to grab the rope during action 2.
This was also true for most failure cases in other experiments.

We visualized the hidden states of the LSTM by Principal Component Analysis (PCA), as depicted in Fig.~\ref{fig:pca1-1}.
The plotted point sequences denote successful examples of each knotting operation.
The orange points correspond to bowknot operations, and the blue points correspond to overhand knot operations.
The initial and final points overlapped with a bowknot and an overhand knot, but the plots were observed to separate gradually as the task progressed.
The most divided part of sequences on the upper left of Fig.~\ref{fig:pca1-1} corresponded to action 2.
The point clouds of the overhand knot and the bow knot were separated only in this section, and the model was observed to successfully generate the appropriate motion based on the sensor information.

\subsection{Experiment 2: Generating Motion Suitable for Rope State}
\label{sec:result2}

Next, we verified the capability of the proposed method to generate the appropriate motion corresponding to the rope state.
The model was trained using the task sequences of overhand knot operations based on three patterns of hooking the rope to the gripper.
Each pattern was tested 20 times using online motion generation.
The success rates of the hand-crafted knotting operation and the proposed method are recorded in the first and second rows of Table III.
The hand-crafted knotting operations indicate the method used to construct the training dataset, and their success rates were counted during the collection of 15 training sequences.
The average success rate of the proposed method was 96.7 \%, which corresponded to a sufficiently high accuracy for the generated motion.
In contrast, the success rate of the hand-crafted knotting operations was quite low.
This was attributed to the change in the rope state depending on the robot's action.
The robot could accomplish tasks only when the pattern of the rope state matched the program of the hand-crafted knotting operations.
The proposed method was verified to be effective because with the accumulation of greater amounts of task experiences, the experience was more generalized and the task accuracy was improved.

We visualized the hidden state of the LSTM by PCA, as depicted in Fig.~\ref{fig:pca1-2}.
The plotted point sequences denote successful examples of each knotting operation.
The blue points correspond to pattern 1 of the rope state as illustrated in Fig.\ref{fig:rope-kakari}, the orange points correspond to pattern 2, and the green points correspond to pattern 3.
The most divided part of sequences on the bottom left of Fig.~\ref{fig:pca1-2} corresponded to action 2.
As in the case of Fig.~\ref{fig:pca1-1}, it is evident that LSTM successfully predicts the appropriate operation based on the sensor information in this action part.
Thus, combined with the result described in Subsection~\ref{sec:result1}, it can be concluded that the proposed method is capable of autonomously executing the motions learned on the basis of sensor information even if the operation to be performed is not explicitly indicated.

\subsection{Experiment 3: Performing Task with Untrained Rope}
\label{sec:result3}

Finally, we evaluated the generalization characteristics of the proposed method by performing tasks with an unlearned rope.
The success rates of the bowknot and overhand knot operations with the unlearned rope are presented in the second row of Table II.
Figure~\ref{fig:result} depicts examples of the generated knotting motions using the unlearned rope.
The success rates were observed to be 85.0 \% for overhand knots and 90.0 \% for bowknots.
Although the success rates were slightly lower than the corresponding ones for the learned rope, they were sufficient to establish that the robot was capable of performing tasks even with a different colored rope.

The third row of Table III presents the success rates corresponding to the different patterns of hooking the rope to the gripper in the overhand knot operation with an unlearned rope.
The average success rate of the proposed method was observed to be 88.3 \%, which corresponded to a sufficiently high accuracy of the generated motion.
The robot had to search for the rope while grabbing it, and the model was observed to treat the information obtained from the proximity sensor to be critical to action 2.
Detailed behaviors of the robot can be observed in the supplementary video.
In our experiments, the system performed tasks at a slow speed for safety. However, a faster operation is also possible by improving the sensor acquisition unit and robot control unit.
Based on the aforementioned results, it can be concluded that the proposed method is capable of predicting the robot motion by successfully generalizing the color of the rope observed using the visual sensor, the state of the rope, and the information obtained from the proximity sensor.

We visualized the averages of the hidden state in the generated sequences by PCA.
Figure~\ref{fig:pca2-1} depicts the results of 5 trials for each knotting task in experiments 1 and 3.
Further, Fig.~\ref{fig:pca2-2} presents the results of 5 trials for each pattern of hooking the rope to the gripper in the overhand knot operation in experiments 2 and 3.
The plotted points were grouped corresponding to each condition, and it is evident that the characteristics of each operation were appropriately embedded in the LSTM.


\section{Conclusion}

In this study, we executed in-air knotting of rope using a dual-arm two-fingers robot based on deep learning.
By training CAE and LSTM using the sensorimotor experiences of the robot, including the information obtained via visual and proximity sensors, we constructed a model enabling the robot to perform the required tasks by dynamically responding to the state of the rope.
The task motions were designed based on the Ian knot method to avoid the requirement of dedicated hardware.
In the verification experiments, the robot was required to perform overhand knot and bowknot operations.
The results confirmed that the robot was capable of autonomously generating appropriate motions online corresponding to all rope states.
In addition, the proposed method generalized the sensorimotor experiences of the robot and could handle unlearned objects.
In future works, we intend to expand the proposed method by including a torque sensor to eliminate looseness after binding the rope.


\section*{Acknowledgment}
This work was based on results obtained from a project, JPNP20006, commissioned by the New Energy and Industrial Technology Development Organization (NEDO).
And also, this work was supported by JST, ACT-X Grant Number JPMJAX190I, Japan.

\end{document}